\documentclass[10pt,twocolumn,letterpaper]{article}

\usepackage[pagenumbers]{cvpr} 

\usepackage[dvipsnames]{xcolor}

\usepackage{colortbl}
\usepackage{adjustbox}

\usepackage{pifont}
\newcommand{\cmark}{\ding{51}}%
\newcommand{\xmark}{\ding{55}}%

\definecolor{inchworm}{rgb}{0.7, 0.93, 0.36}
\definecolor{junebud}{rgb}{0.74, 0.85, 0.34}
\definecolor{lightpastelpurple}{rgb}{0.69, 0.61, 0.85}
\newcommand{\ourcolor}[1]{\cellcolor{junebud!25} #1}

\definecolor{deepblue}{rgb}{0,0,0.5}
\definecolor{deepred}{rgb}{0.6,0,0}
\definecolor{deepgreen}{rgb}{0,0.5,0}

\definecolor{codegreen}{rgb}{0,0.6,0}
\definecolor{codegray}{rgb}{0.5,0.5,0.5}
\definecolor{codepurple}{rgb}{0.58,0,0.82}
\definecolor{backcolour}{rgb}{0.95,0.95,0.92}
\definecolor{light-gray}{gray}{0.8}

\usepackage[ruled]{algorithm2e}
\usepackage{setspace}

\usepackage{listings}

\definecolor{codeblue}{rgb}{0.25,0.5,0.5}
\newcommand{\tablestyle}[2]{\setlength{\tabcolsep}{#1}\renewcommand{\arraystretch}{#2}\centering\footnotesize}
\newlength\savewidth\newcommand\shline{\noalign{\global\savewidth\arrayrulewidth
		\global\arrayrulewidth 1pt}\hline\noalign{\global\arrayrulewidth\savewidth}}
\usepackage{tabulary,multirow,overpic}

\newcommand{\blockatt}[3]{\multirow{2}{*}{\(\left[\begin{array}{c}\text{MHA(\wcolor{#1})}\\[-.1em] \text{MLP(\wcolor{#2})}\end{array}\right]\)$\times$#3}
}
\newcommand{\blockxatt}[3]{\multirow{3}{*}{\(\left[\begin{array}{c}\text{MHCA(\wcolor{#1})}\\[-.1em] 
\text{MHA(\wcolor{#1})}\\[-.1em] 
\text{MLP(\wcolor{#2})}\end{array}\right]\)$\times$#3}
}

\definecolor{linkcol}{RGB}{233, 4, 141}

\definecolor{xycolor}{RGB}{60, 120, 216}
\definecolor{xycolor}{HTML}{0071bc}
\newcommand{\xycolor}[1]{\textcolor{xycolor}{#1}}
\definecolor{wcolor}{RGB}{103, 78, 167}
\newcommand{\wcolor}[1]{\textcolor{wcolor}{#1}}
\definecolor{dcolor}{RGB}{166, 77,21}

\definecolor{gcolor}{RGB}{204, 102, 153}

\definecolor{tcolor}{RGB}{34,139,34}
\newcommand{\tcolor}[1]{\textcolor{tcolor}{#1}}
\definecolor{iterc}{RGB}{91,196,159}

\definecolor{epochc}{RGB}{96,172,252}

\definecolor{eicolor}{RGB}{153, 51, 102}

\newcommand{\maskcolor}[1]{\textcolor{orange}{#1}}

\def\x{$\times$}

\newcolumntype{x}[1]{>{\centering\arraybackslash}p{#1pt}}
\newcolumntype{y}[1]{>{\raggedright\arraybackslash}p{#1pt}}
\newcolumntype{z}[1]{>{\raggedleft\arraybackslash}p{#1pt}}

\definecolor{cvprblue}{rgb}{0.21,0.49,0.74}
\usepackage{hyperref}
\hypersetup{breaklinks,colorlinks,citecolor=cvprblue}

\def\confName{CVPR}
\def\confYear{2024}

\title{MV2MAE: Multi-View Video Masked Autoencoders}

\author{
    Ketul Shah\textsuperscript{1} ~~~~
    Robert Crandall\textsuperscript{2} ~~~~
    Jie Xu\textsuperscript{2} ~~~~
    Peng Zhou\textsuperscript{2} ~~~~
    Marian George\textsuperscript{2} ~~~~\\
    Mayank Bansal\textsuperscript{2} ~~~~
    Rama Chellappa\textsuperscript{1} ~~~~
	\\
	\textsuperscript{1}Johns Hopkins University \ \ \ \ \ 
	\textsuperscript{2}Amazon  \ \ \ \ \  \\
	{\small \tt \{kshah33, rchella4\}@jhu.edu~\{rcrandal, xumji, zhoupz, margeo, maybans\}@amazon.com}
}

\begin{document}
\maketitle

\begin{abstract}
    Videos captured from multiple viewpoints can help in perceiving the 3D structure of the world and benefit computer vision tasks such as action recognition, tracking, etc. In this paper, we present a method for self-supervised learning from synchronized multi-view videos. We use a cross-view reconstruction task to inject geometry information in the model. Our approach is based on the masked autoencoder (MAE) framework. In addition to the same-view decoder, we introduce a separate cross-view decoder which leverages cross-attention mechanism to reconstruct a target viewpoint video using a video from source viewpoint. This helps learn representations robust to viewpoint changes. For videos, static regions can be reconstructed trivially which hinders learning meaningful representations. To tackle this, we introduce a motion-weighted reconstruction loss which improves temporal modeling. We report state-of-the-art results on the NTU-60, NTU-120 and ETRI datasets, as well as in the transfer learning setting on NUCLA, PKU-MMD-II and ROCOG-v2 datasets, demonstrating the robustness of our approach. Code will be made available.  
\end{abstract}

\section{Introduction}

Multiple viewpoints of the same event are crucial to its understanding. Humans move around and obtain different viewpoints of objects and scenes, and develop a representation robust to viewpoint changes~\cite{isik2018fast}. Different viewpoints often have very different appearance, which can help address challenges due to occlusion, lighting variations and limited field-of-view. In many real world scenarios, we have videos captured from multiple viewpoints, \eg sports videos~\cite{1394492}, elderly care~\cite{etri}, self-driving~\cite{yogamani2019woodscape}, complex robotic manipulation tasks~\cite{mvmwm} and security videos~\cite{Corona_2021_WACV}. Learning a robust pre-trained model from large amounts of unlabeled synchronised multi-view data is of significant value for these applications. Such a model which is aware of the 3D geometry will be robust to changes in viewpoint and can be effectively used as a foundation for downstream finetuning on smaller datasets for different tasks. 

There has been significant progress in video self-supervised learning~\cite{schiappa2023self} for the single-view case, \ie where synchronized multi-view data is not available. Recently, Masked Autoencoders (MAEs) as a paradigm for self-supervised learning has seen growing interest, and it has been successfully extended to video domain ~\cite{maeST, videomae, videomaev2}.
MAE-based methods achieve superior performance~\cite{videomae} on standard datasets such as Kinetics-400 ~\cite{kay2017kinetics} and Something-Something-v2 ~\cite{goyal2017something}, compared to contrastive learning methods~\cite{Feichtenhofer2021ALS}. 
However, existing MAE-based pre-training approaches are not explicitly designed to be robust to viewpoint changes. View-invariant learning from multi-view videos has been widely studied using NTU~\cite{ntu60, ntu120} and ETRI~\cite{etri} datasets. However, most of these methods are based on 3D human pose, which is difficult to accurately capture or annotate for in-the-wild scenarios. Recently, there has been a growing interest in RGB-based self-supervised learning approaches leveraging multi-view videos~\cite{parameswaran2006view, viewclr, vyas2020multi, li2018unsupervised}, facilitated by the availability of large-scale multi-view datasets~\cite{ntu60, ntu120, etri}. ViewCLR~\cite{viewclr}, which achieves state-of-the-art results, introduces a latent viewpoint generator as a learnable augmentation for generating positives in a contrastive learning~\cite{chen2020improved} framework. However, this method is memory intensive as it requires storing two copies of the feature extractor and two queues of features, and also requires multi-stage training. Considering the recent success of MAEs for video SSL, it is desirable to explore its potential in the multi-view video SSL scenario.  

\begin{figure*}[t]
\begin{center}
   \includegraphics[scale=0.52]{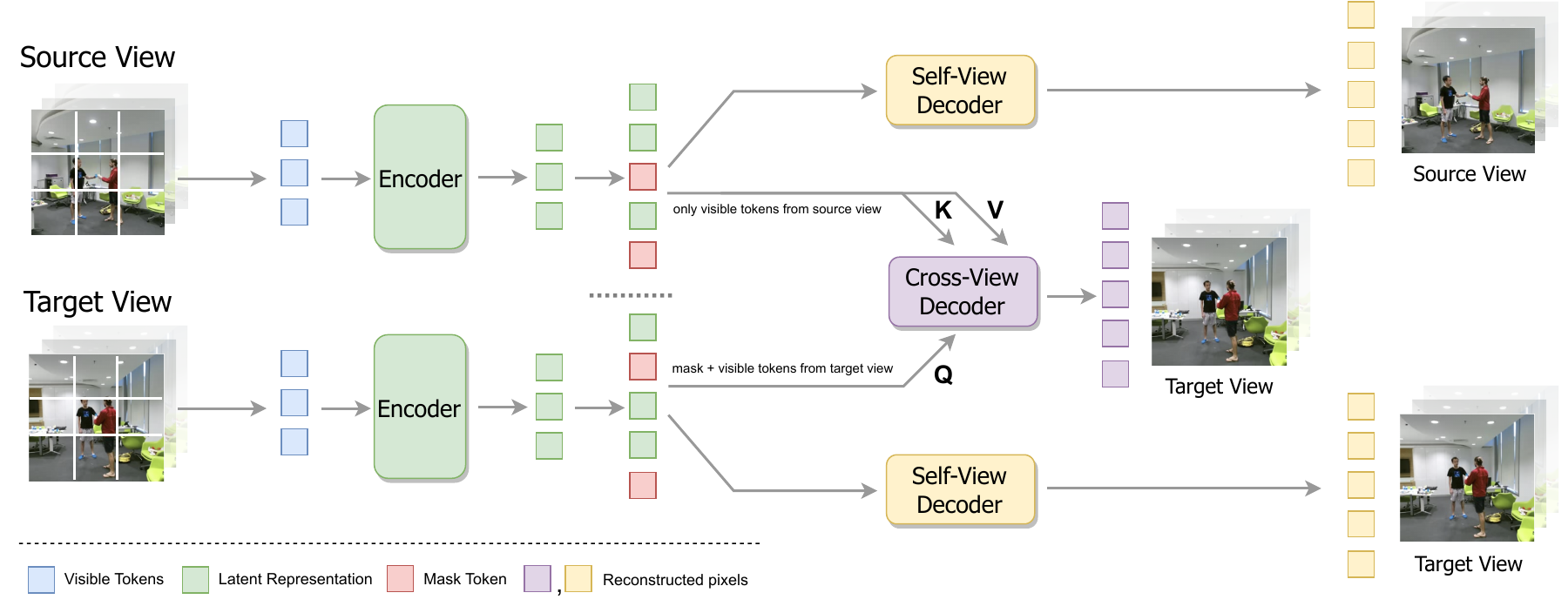}
\end{center}
   \caption{\textbf{Multi-View Video Masked Autoencoder (MV2MAE).} Source and target viewpoint videos are tokenized, the tokens are masked using random masking, and visible tokens are encoded for each view using a shared encoder. The cross-view decoder uses source view tokens to reconstruct the target viewpoint, whereas the standard decoder reconstructs each view separately.}
\label{fig:overview}
\vspace{-0.2cm}
\end{figure*}

In this paper, we aim to learn a self-supervised video representation which is robust to viewpoint shifts. Humans learn a view invariant representation for tasks such as action recognition and are able to \textit{visualize how an action looks from different viewpoints}~\cite{isik2018fast}. We integrate this task of using features of one viewpoint to predict the appearance of a video from a different viewpoint in the standard MAE framework. More specifically, given a video of an activity from one viewpoint, it is patchified and a high fraction of the patches are masked out. The visible patches are encoded, which the decoder uses (along with \texttt{MASK} tokens for missing patches) to reconstruct the given video. We introduce an additional cross-view decoder, which is tasked with reconstructing the masked patches of a target viewpoint by using the visible regions of from source view. This would require the model to understand the geometric relations between different views, enabling the construction of a robust pre-trained model. A challenge with MAE in videos is that they contain a lot of temporal redundancy, making it easier to reconstruct the static, background regions by simply copy pasting from adjacent frames where those are visible. Existing solutions for this problem involve specialized masking strategies using extra learnable modules ~\cite{adamae, mgmae} or use tube masking~\cite{videomae, videomaev2}, which are not effective in certain scenarios, \eg when motion is localized in a small region of the frame. We propose a simple solution without introducing additional learnable parameters by modifying the reconstruction loss to focus on moving regions. We can control the relative weights of moving and static regions using a temperature parameter.

We perform experiments on three multi-view video datasets: NTU-60~\cite{ntu60}, NTU-120~\cite{ntu120}, ETRI~\cite{etri} for pre-training. Our method achieves SOTA fine-tuning accuracy on all benchmarks of these datasets. The robustness of our representation is shown in the transfer learning results on smaller datasets. We achieve SOTA results on NUCLA~\cite{wang2014cross}, ROCOG-v2~\cite{2023rocogv2} and PKU-MMD-II~\cite{liu2017pku} datasets in the transfer learning setting. 

\medskip

Our main contributions can be summarized as follows:
\medskip
\begin{itemize}
    \setlength\itemsep{5pt}
    \item We present an approach for self-supervised pre-training from multi-view videos using the MAE framework, and achieve state-of-the-art results on a variety on benchmarks under full finetuning and transfer learning settings. 
    \item Our approach uses cross-view reconstruction to inject geometry information in the model. This is done via a separate decoder with cross-attention mechanism which reconstructs target view from source view. 
    \item We introduce a simple motion-focused reconstruction loss for improved temporal modeling, while also allowing us to specify the degree to which to focus on moving regions. 
\end{itemize}

\section{Related Work}

\subsection{Self-Supervised Learning from Videos}
\noindent\textbf{Pretext Learning.}
Many pretext tasks have been proposed for learning self-supervised video representations, initially inspired from the progress in SSL for images. Tasks such as video rotation prediction~\cite{jing2018self}, solving spatio-temporal jigsaw~\cite{ahsan2019video}, predicting motion and appearance statistics~\cite{wang2019self} were direct extensions of their image counterparts, and showed impressive performance. Methods leveraging the temporal order in videos for constructing pretext tasks such as frame ordering~\cite{xu2019self} and odd-one-out learning~\cite{fernando2017self} were also proposed. These methods were outperformed by contrastive learning approaches. 

\noindent\textbf{Contrastive Learning.}
These methods create augmented versions of the input (positives) which preserve the semantic content of the input. The contrastive loss is used to pull these closer together in the feature space, while simultaneously pushing them away from other samples (negatives). Numerous ways of generating positive pairs were proposed such as using random clips from the same video, clips of different frame rates~\cite{wang2020self}, choosing nearby clips~\cite{qian2021spatiotemporal}, and using optical flow~\cite{coclr}, among others. 

\noindent\textbf{Masked Video Modeling.}
Recently, masked video modeling has emerged as a promising area for self-supervised learning. Methods such as BEVT~\cite{bevt}, MaskFeat~\cite{maskfeat}, VideoMAE~\cite{videomae}, MAE-ST~\cite{maeST} show superior performance on the standard video self-supervised learning benchmarks. Different reconstruction targets have been studied, such as MVD~\cite{mvd} which uses distillation from pre-trained features, and MME~\cite{mme} which reconstructs motion trajectories. To tackle trivial reconstruction solution via copy-paste in videos, which becomes and issue due to high redundancy, different masking strategies have been proposed. MGMAE~\cite{mgmae} uses motion-guided masking based on motion vectors, VideoMAE~\cite{videomae} proposed using tube masking, AdaMAE~\cite{adamae} introduces a neural network for mask sampling. Orthogonal to these, we propose to tackle the issue by using a motion-weighted reconstruction loss. Moreover, unlike our approach, existing MAE pre-training approaches are not explicitly designed to be robust to viewpoint shifts.   

\subsection{Multi-View Action Recognition}
Early work in this area designed hand-crafted features which were robust to viewpoint shifts~\cite{parameswaran2006view, rao2002view, xia2012view}. Many unsupervised learning approaches have been proposed for learning representations robust to changes in viewpoint. A large number of methods leverage 3D human pose information, which greatly aids in achieving view invariance. Methods based on RGB modality~\cite{viewclr, li2018unsupervised, vyas2020multi} have gained increasing popularity. These can be broadly divided into two categories:   

One trend is to enforce the latent representations of different viewpoints to be close. Along this line,~\cite{zheng2012cross} follows a dictionary learning approach and encourages videos of different views to have the same sparse representation.~\cite{rahmani2015learning} fits a 3D human model to a mocap sequence and generates videos from multiple viewpoints, which are forced to predict the same label. More recently, methods based on contrastive learning have been proposed such as ViewCLR~\cite{viewclr} which achieves remarkable performance. They add a latent viewpoint generator module which is used to generate positives in the latent space corresponding to different views. 

Another line of work uses one viewpoint to predict another. ~\cite{li2018unsupervised} uses cross-view prediction in 3D flow space by using depth as an additional input to provide view information. Their approach also uses a gradient reversal layer for enforcing view invariance.~\cite{vyas2020multi} uses the encoded source view features to render same video from unseen viewpoint and a random start time. Their approach hence needs to be able to predict across time and viewpoint shifts. They leverage a view embedding which requires information of camera height, distance and angle. In contrast the these approaches which rely on view embedding or depth for providing viewpoint information, the view information is inherently available in the visible patches of the viewpoints in our approach.

\section{Method}

\subsection{Preliminary: Masked Video Modeling}
Here we revisit the masked autoencoder (MAE) framework for videos. Given a video, we first sample $T$ frames with stride $\tau$ to get the input clip: $\textbf{I} \in \mathbb{R}^{C \times T \times H \times W}$. Here, $H \times W$ is the spatial resolution, $T$ denotes the number of frames sampled, and $C$ is the number of input (RGB) channels. The standard MAE architecture has three main components: tokenizer, encoder, decoder. 

\noindent\textbf{Tokenizer.} The input clip is first converted into $N$ patches using a patch size of $t \times h \times w$, where $N = \frac{T}{t} \times \frac{H}{h} \times \frac{W}{w}$. The tokenizer returns $N$ tokens of dimension $d$ by first linearly embedding these $N$ patches. This is implemented in practice using a strided 3D convolution layer. Next, we provide position information to these tokens by adding positional embeddings ~\cite{vaswani2017attention}.  

\noindent\textbf{Encoder.} A high fraction of these $N$ tokens are dropped with a masking ratio $\rho \in (0,1)$. Different masking strategies~\cite{videomae, adamae, mgmae} have been explored for choosing which tokens to mask out. Next, the remaining small fraction of visible tokens are passed through the encoder ($\Phi_{\texttt{enc}}$) to obtain latent representations. The encoder is a vanilla ViT~\cite{dosovitskiy2020image} with joint space-time attention~\cite{videomae}. These latent representations need to capture the semantics in order to reconstruct the masked patches.

\noindent\textbf{Decoder.} The encoded latent representations of the visible patches are concatenated with learnable $\texttt{MASK}$ tokens corresponding to masked out patches, resulting in combined tokens $\textbf{Z}_{\texttt{c}}$. Then the positional embeddings are added for all tokens, and passed through a light-weight decoder ($\Phi_{\texttt{dec}}$) to get the predicted pixel values $\hat{\textbf{I}} = \Phi_{\texttt{dec}}(\textbf{Z}_{\texttt{c}})$.  

The loss function is the mean squared error (MSE) between the reconstructed values and the normalized pixel values~\cite{maeST, videomae}, for masked patches $\Omega$. 

\begin{equation}
    \mathcal{L} = \frac{1}{\rho N} \sum_{i \in \Omega} | \textbf{I}_i - \hat{\textbf{I}}_i |^2
\end{equation}

\subsection{Cross-View Reconstruction}
The goal of cross-view reconstruction is to predict the missing appearance of a video in target viewpoint given videos from one (few) source viewpoint(s). Being able to extrapolate across viewpoints requires understanding the geometric relations between different viewpoints, making it an effective task for learning representations robust to viewpoint variations. 

As shown in Figure~\ref{fig:overview}, consider two synchronized videos of an activity, $\textbf{I}^{\texttt{sv}}$ and $\textbf{I}^{\texttt{tv}}$, from source view ($\texttt{sv}$) and target view ($\texttt{tv}$) respectively. We first tokenize, mask and encode the visible tokens for each video separately using a shared encoder $\Phi_{\texttt{enc}}$. 
We introduce a cross-view decoder ($\Phi_{\texttt{dec}}^{\texttt{cross-view}}$) which uses a cross-attention mechanism. This decoder additionally uses the visible tokens in the source view to reconstruct the target viewpoint video, $\hat{\textbf{I}}^{\texttt{tv}} = \Phi^{\texttt{cross-view}}_{\texttt{dec}}(\textbf{Z}_{\texttt{c}}^{\texttt{tv}}, \textbf{Z}_{\texttt{vis}}^{\texttt{sv}})$.
More specifically, each block of the cross-view decoder consists of cross-attention and self-attention layers, followed by a feed-forward layer. The tokens from the target view attend to the visible source view tokens using cross-attention, and then to each other using self-attention. A key aspect of methods based on cross-view prediction is how the viewpoint information is provided:~\cite{vyas2020multi} conditions the decoder on a viewpoint embedding, while some approaches~\cite{li2018unsupervised} use extra modalities such as depth to provide information about the target viewpoint. In contrast to these, the visible patches provide the required target viewpoint information in our approach. The \textit{amount} of view information we want to provide can be easily varied by changing the masking ratio. 
Moreover, the standard decoder ($\Phi_{\texttt{dec}}$) is used to reconstruct video from each viewpoint independently $\hat{\textbf{I}}^{\texttt{vp}} = \Phi_{\texttt{dec}}(\textbf{Z}_{\texttt{c}}^{\texttt{vp}})$ for $ \texttt{vp} \in \{\texttt{sv}, \texttt{tv}\}$. 

Figure~\ref{fig:xattn} visualizes the cross-view reconstruction quality and cross-attention maps, which demonstrates that the model learns to focus on relevant regions across viewpoints.

\begin{figure}[t]
\begin{center}
   \includegraphics[scale=0.45]{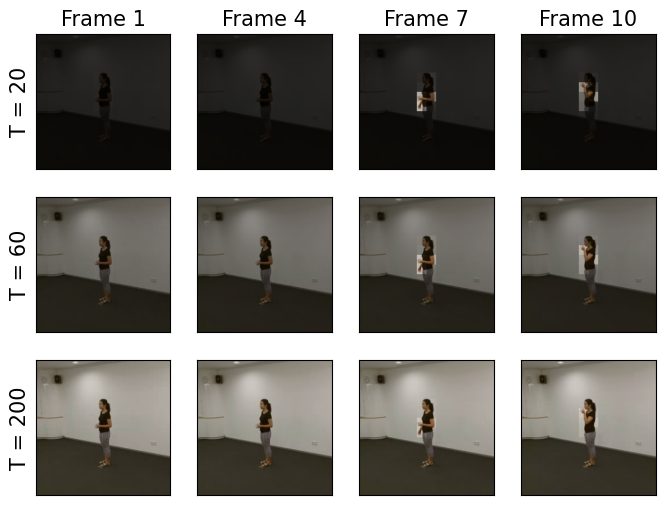}
    
\end{center}
   \caption{\textbf{Motion weights.} Each row shows motion weights overlaid on the input frames for different temperature values. Higher temperature increases the weight on static regions.}
\label{fig:motion-weights-examples}
\vspace{-0.2cm}
\end{figure}
\subsection{Motion-Weighted Reconstruction Loss}
A given video can be decomposed into static and dynamic regions. Static regions typically involve scene background and objects which do not move throughout the video. Patches from such regions are trivial to reconstruct~\cite{mme, adamae} due to temporal redundancy in videos. In order to deal with this, we offer a simple solution by re-weighting the reconstruction loss of each patch proportional to the amount of motion within that patch. The motion weights used for re-weighting are obtained using frame difference for simplicity. Note that other motion features such as optical flow, motion history image, etc can be used in place of frame difference, but frame difference is extremely fast to compute. In order to get the final weights, we take the norm of frame difference within each patch, and apply softmax over all tokens, as shown in Algorithm~\ref{code}. We can control the extent to which to focus on the moving regions by controlling the temperature parameter. The higher the temperature value, the more uniform the resulting weights. Examples of motion weights overlaid on the original frames for different temperature values are shown in Figure~\ref{fig:motion-weights-examples}. PyTorch-style code for computing the motion weights for a video is presented in Algorithm~\ref{code}. The final motion-weighted reconstruction loss is given below, where $w_i$ is the weight for $i^{th}$ patch:   
\begin{equation}
    \mathcal{L} = \frac{1}{\rho N} \sum_{i \in \Omega} w_i \times | \textbf{I}_i - \hat{\textbf{I}}_i |^2
\end{equation}

\newcommand\algcomment[1]{\def\@algcomment{\footnotesize#1}}
\begin{algorithm}[t]
\caption{PyTorch code for motion weights.}\label{code}
\begin{lstlisting}[language=python]
# frames    : input frames of shape [B,C,T,H,W]
# patch_size: (p_time, p_height, p_width)
# t         : temperature parameter

fdiff = frames[:,:,1:,:,:] - frames[:,:,:-1,:,:]
fdiff = torch.cat([fdiff[:,:,0:1,:,:], fdiff], dim=2)
fdiff = rearrange(fdiff,'b c (t p0) (h p1) (w p2) -> b (t h w) (p0 p1 p2 c)', p0=patch_size[0], p1=patch_size[1], p2=patch_size[2])
fdiff = torch.abs(fdiff)
fdiff = torch.linalg.vector_norm(fdiff, dim=2, keepdim=True) # B N 1 
motion_weights = torch.nn.functional.softmax(fdiff/t, dim=1) # B N 1 
\end{lstlisting}
\end{algorithm}
\section{Experiments}
We evaluate our approach on several common multi-view video datasets: NTU60~\cite{ntu60}, NTU120~\cite{ntu120}, ETRI~\cite{etri}, NUCLA~\cite{wang2014cross}, PKU-MMD~\cite{liu2017pku}, and ROCOG \cite{2023rocogv2}. For NTU and ETRI, we achieve state-of-the-art results by pre-training and fine-tuning on the target domain. On NUCLA, PKU-MMD-II, and ROCOG-v2, we demonstrate excellent transfer learning performance by pre-training only on NTU, and fine-tuning on the target dataset.

\subsection{Datasets}
\noindent\textbf{NTU RGB+D 60.} ~\cite{ntu60} is a large-scale multi-view action recognition dataset, consisting of 56,880 videos from 60 distinct action classes. These videos were recorded from 40 subjects using Kinect-v2. Each activity instance is simultaneously captured from three viewpoints. The dataset consists of two benchmarks outlined in~\cite{ntu60}: (1) Cross-Subject (xsub) and (2) Cross-View (xview). In the cross-subject benchmark, the 40 subjects are divided into training and testing sets, with 20 subjects in each. In the cross-view scenario, videos from cameras 2 and 3 are used for training, while testing is performed on videos from camera 1.

\vspace{5pt}
\noindent\textbf{NTU RGB+D 120.} \cite{ntu120} is the extended version of the NTU-60 dataset which contains 114,480 videos spanning 120 action categories. Our evaluation follows the established protocols outlined in \cite{ntu120}: (1) Cross-Subject (xsub) and (2) Cross-Setup (xset). In the cross-subject scenario, subjects are partitioned into training and testing groups, while in the cross-setup setting, the data is divided into training and testing subsets based on the setup ID.

\vspace{5pt}
\noindent\textbf{ETRI.} ~\cite{etri} is another large-scale multi-view action recognition dataset consisting of activities of daily living for elderly care. It has 112,620 videos captured from 55 action classes. All activity instances are recorded from 8 synchronized viewpoints. ~\cite{etri} describes a cross-subject benchmark which we use to evaluate our approach.  



\begin{table}[t]

\caption{Comparison with state-of-the-art on cross-view and cross-subject benchmarks of NTU-60 dataset. \textbf{Top:} supervised methods using multiple modalities, \textbf{Middle:} supervised methods using only RGB modality, \textbf{Bottom:} unsupervised methods using any modality. Labels \cmark : Supervised methods}

  \begin{center}
    {\small{
\begin{tabular}{lcccc}
\toprule
& & & \multicolumn{2}{c}{\textbf{NTU-60 (\%)}} \\
Method & Modality & Labels & xview & xsub \\
\midrule

STA-Hands \cite{baradel2017human} & RGB+Pose & \cmark & 88.6 & 82.5 \\
Separable STA \cite{Das_2019_ICCV} & RGB+Pose & \cmark & 94.6 & 92.2 \\
VPN \cite{das2020vpn} & RGB+Pose & \cmark & 96.2 & 93.5 \\

ESE-FN~\cite{shu2022expansion} & RGB+Pose & \cmark & 96.7 & 92.4 \\ 

\midrule
DA-Net~\cite{wang2018dividing} & RGB & \cmark & 75.3 & -- \\
Zhang \etal~\cite{zhang2018adding} & RGB & \cmark & 70.6 & 63.3 \\
Glimpse Clouds~\cite{Baradel_2018_CVPR} & RGB & \cmark & 93.2 & 86.6 \\
DMCL~\cite{garcia2021distillation} & RGB & \cmark & -- & 83.6 \\ 
Debnath \etal~\cite{debnath2021attentional} & RGB & \cmark & -- & 87.2 \\
FSA-CNN~\cite{etri} & RGB & \cmark & 92.2 & 88.1 \\ 
Piergiovanni \etal~\cite{Piergiovanni_2021_CVPR} & RGB & \cmark & 93.7 & -- \\
ViewCon~\cite{viewcon} & RGB & \cmark & 98.0 & 91.4 \\

\midrule
Li \etal~\cite{li2018unsupervised} & Flow & \xmark & 83.4 & 80.9 \\ 
\arrayrulecolor{light-gray}
\midrule
HaLP~\cite{shah2023halp} & Pose & \xmark & 88.6  & 82.1 \\ 
\midrule
Vyas \etal \cite{vyas2020multi} & RGB & \xmark & 86.3 & 82.3 \\
ViewCLR~\cite{viewclr}  & RGB & \xmark & 94.1  & 89.7 \\

\arrayrulecolor{black}
\midrule
\ourcolor MV2MAE (Ours) & \ourcolor RGB & \ourcolor \xmark & \ourcolor \textbf{95.9} & \ourcolor \textbf{90.0} \\

\bottomrule
\end{tabular}
}}
\\
\end{center}
\label{table:ntu60}
\end{table}
\begin{table}[t]

\caption{Comparison with state-of-the-art on cross-setup and cross-subject benchmarks of NTU-120 dataset. \textbf{Top:} supervised methods using multiple modalities, \textbf{Middle:} supervised methods using single modality, \textbf{Bottom:} unsupervised methods. MV2MAE outperforms previous SOTA unsupervised methods using any modality. Labels \cmark : Supervised methods}

  \begin{center}
    {\small{
\begin{tabular}{lcccc}
\toprule
 & & & \multicolumn{2}{c}{\textbf{NTU-120 (\%)}} \\
Method & Modality & Labels & xset & xsub \\
\midrule

Hu \etal~\cite{hu2018early} & RGB+Depth & \cmark & 44.9 & 36.3 \\
Hu \etal~\cite{hu2015jointly} & RGB+Depth  & \cmark & 54.7 & 50.8 \\
Separable STA~\cite{Das_2019_ICCV} & RGB+Pose & \cmark & 82.5 & 83.8 \\ 
VPN~\cite{das2020vpn} & RGB+Pose & \cmark & 87.8 & 86.3 \\

\midrule
\arrayrulecolor{light-gray}
PEM~\cite{pem} & Pose & \cmark & 66.9 & 64.6 \\
2s-AGCN~\cite{2sagcn2019cvpr} & Pose & \cmark & 84.9 & 82.9 \\
MS-G3D Net~\cite{msg3d} & Pose & \cmark & 88.4 & 86.9 \\
CTR-GCN~\cite{ctr_gcn} & Pose & \cmark & 90.6 & 88.9 \\
\midrule 
Two-streams~\cite{twostream} & RGB & \cmark & 54.8 & 58.5 \\ 
Liu \etal~\cite{ntu120} & RGB  & \cmark & 54.8 & 58.5 \\
I3D~\cite{i3d} & RGB & \cmark & 80.1 & 77.0 \\ 
DMCL~\cite{garcia2021distillation} & RGB & \cmark & 84.3 & -- \\
ViewCon~\cite{viewcon} & RGB & \cmark & 87.5 & 85.6 \\ 

\arrayrulecolor{black}
\midrule
\arrayrulecolor{light-gray}
HaLP~\cite{shah2023halp} & Pose & \xmark & 73.1 & 72.6 \\
\midrule
ViewCLR~\cite{viewclr}  & RGB & \xmark & 86.2 & 84.5 \\
\arrayrulecolor{black}
\midrule
\ourcolor MV2MAE (Ours) & \ourcolor RGB & \ourcolor \xmark & \ourcolor \textbf{87.1} & \ourcolor \textbf{85.3} \\

\bottomrule
\end{tabular}
}}
\end{center}
\label{table:ntu120}
\end{table}

\begin{table}[ht]

\caption{Comparison with state-of-the-art cross-subject benchmark of ETRI dataset. MV2MAE performs better than prior work, which are all supervised approaches.}
  \begin{center}
    {\small{
\begin{tabular}{lccc}
\toprule
 & & & \textbf{ETRI (\%)} \\
Method & Modality & Labels & xsub \\
\midrule
ESE-FN~\cite{shu2022expansion} & RGB+Pose & \cmark & 95.9 \\
FSA-CNN~\cite{etri} & RGB & \cmark & 90.6 \\ 
ConViViT~\cite{convivit} & RGB & \cmark & 95.1 \\ 

\midrule

\ourcolor MV2MAE (Ours) & \ourcolor RGB & \ourcolor \xmark & \ourcolor \textbf{96.5} \\

\bottomrule
\end{tabular}
}}
\end{center}
\label{table:etri}
\end{table}

\subsection{Implementation Details}
We sample a clip of 16 RGB frames with a stride of 4 from each video. We downsample the resolution of frames to $128 \times 128$ following ~\cite{viewclr}. During pre-training, we only apply random resized crops as augmentation. We use a temporal patch size of $2$ and a spatial patch size of $16 \times 16$, which results in $512$ tokens. A masking ratio of 0.7 is used unless otherwise specified. We choose fixed sinusoidal spatio-temporal positional position embedding following~\cite{videomae, adamae}. All of our experiments use the vanilla ViT-S/16~\cite{touvron2021training} architecture as the encoder (unless otherwise noted), trained using AdamW optimizer~\cite{loshchilov2017decoupled}. The pre-training is carried out for 1600 epochs. Please refer to the supplementary material for more details. 

We evaluate our pre-trained models using two settings: 1) end-to-end fine-tuning on the same datasets and 2) transfer learning on smaller datasets. 
We discard the decoders and attach a classifier head which uses the global average pooled features for classification. For testing, we sample $5$ temporal clips, and use $10$ crops from each following~\cite{viewclr}, and the final prediction is the average of these.

\begin{table}[t]

\caption{Transfer learning on NUCLA. Unsupervised methods (Labels: \xmark) have been pre-trained on NTU-60 dataset. MV2MAE significantly outperforms other methods showing remarkable transfer capability of our representations.}

  \begin{center}
    {\small{
\begin{tabular}{lccc}
\toprule
 & & & \textbf{NUCLA (\%)} \\
Method & Modality & Labels & xview \\
\midrule
STA~\cite{Das_2019_ICCV} & RGB+Pose & \cmark & 92.4 \\ 
VPN~\cite{das2020vpn} & RGB+Pose & \cmark & 93.5 \\ 
\midrule

DA-Net~\cite{wang2018dividing} & RGB & \cmark & 86.5 \\
Glimpse Cloud~\cite{Baradel_2018_CVPR} & RGB & \cmark & 90.1 \\ 
I3D~\cite{i3d} & RGB & \cmark & 88.8 \\ 
ViewCon~\cite{viewcon} & RGB & \cmark & 91.7 \\

\midrule
\arrayrulecolor{light-gray}
MS\textsuperscript{2}L~\cite{MS2L} & Pose & \xmark & 86.8 \\
\midrule 
Li \etal~\cite{li2018unsupervised} & Depth & \xmark & 62.5 \\
Colorization~\cite{skeleton_colorization_ICCV} & Depth & \xmark & 94.0 \\ 
\midrule
Vyas \etal~\cite{vyas2020multi} & RGB & \xmark & 83.1 \\
ViewCLR~\cite{viewclr} & RGB & \xmark & 89.1 \\

\arrayrulecolor{black}
\midrule

\ourcolor MV2MAE (Ours) & \ourcolor RGB & \ourcolor \xmark & \ourcolor \textbf{97.6} \\

\bottomrule
\end{tabular}
}}
\end{center}
\label{table:nucla}
\end{table}

\subsection{Comparison with state-of-the-art}

We compare our approach with previous SOTA methods on the cross-subject (xsub) and cross-view (xview) benchmarks of the commonly used NTU-60 and NTU-120 datasets. We also present our results on the ETRI dataset, which only has a cross-subject benchmark. 

Table~\ref{table:ntu60} and Table~\ref{table:ntu120} show results on the NTU-60 and NTU-120 datasets. We outperform all previous unsupervised methods based on RGB, Flow or Pose modality on both cross-view and cross-subject benchmarks of the two datasets. On NTU-120, our method approaches the performance of RGB-based \textit{supervised} methods. In the xsub setting, we see an improvement of $+0.3\%$ and $+1.2\%$ on NTU-60 and NTU-120 respectively, and in the xview setting, observe an improvement of $+1.8\%$ and $+0.9\%$ respectively. Our approach is also faster to train~\cite{videomae} and more memory efficient compared to ViewCLR~\cite{viewclr}, which uses a MoCo~\cite{chen2020improved} framework and requires storing two copies of the model and two queues in memory. ~\cite{vyas2020multi} uses a cross-view prediction paradigm but performs poorly ($86.3\%$ vs $95.9\%$ on xview and $82.3\%$ vs $90.0\%$ on xsub) despite using more parameters ($\sim$72M vs $\sim$22M). Unlike their approach which relies only on viewpoint embeddings for information of the target viewpoint, we implicitly have that information in the visible patches from target view, shows the effectiveness of our pre-training mechanism.      

\begin{table}[t]

\caption{Transfer learning on PKU-MMD-II. All methods use NTU-120 dataset for pre-training. MV2MAE surpasses other unsupervised methods, all of which use Pose modality.}

  \begin{center}
    {\small{
\begin{tabular}{lccc}
\toprule
 & & & \textbf{PKU-MMD-II (\%)} \\
Method & Modality & Labels & xsub \\
\midrule

CrosSCLR-B~\cite{zolfaghari2021crossclr} & Pose & \xmark & 52.8 \\ 
CMD~\cite{Mao_2022_CMD} & Pose & \xmark & 57.0 \\ 
HaLP~\cite{shah2023halp} & Pose & \xmark & 57.3 \\ 
\midrule

\ourcolor MV2MAE (Ours) & \ourcolor RGB & \ourcolor \xmark & \ourcolor \textbf{60.1} \\

\bottomrule
\end{tabular}
}}
\end{center}
\label{table:pku}
\end{table}

\begin{table}[t]

\caption{Transfer learning on ROCOG-v2 ground dataset.}
  \begin{center}
    {\small{
\begin{tabular}{lcc}
\toprule
Method & Modality & \textbf{ROCOG-v2 (\%)} \\
\midrule

Reddy \etal~\cite{2023rocogv2} & RGB & 87.0 \\ 
\midrule

\ourcolor MV2MAE (Ours) & \ourcolor RGB & \ourcolor \textbf{89.0} \\
\bottomrule

\end{tabular}
}}
\end{center}
\label{table:rocog}
\end{table}

\subsection{Transfer Learning Results}

Transfer learning is an important setting for evaluating the generalization capabilities of pre-trained models. The model is initialized using pre-trained weights and fine-tuned on smaller datasets. We perform transfer learning experiments on three action recognition datasets: 1) NUCLA, 2) PKU-MMD-II, and 3) ROCOG-v2. 

NUCLA~\cite{wang2014cross} is a multi-view action recognition dataset consisting of 1493 videos spanning 10 action classes. Each activity has been captured from three viewpoints, and we follow the cross-view protocol for our experiments. 
PKU-MMD-II~\cite{liu2017pku} is another dataset for 3D action understanding, consisting of 6945 videos from 51 activity classes. Following prior work~\cite{shah2023halp}, we use the phase 2 of the dataset and evaluate our approach on the cross-subject setting. 
ROCOG-v2~\cite{2023rocogv2} is a gesture recognition dataset consisting of 304 videos from ground viewpoint from 7 gestures.   

As shown in Table~\ref{table:nucla}, our method achieves significantly better performance than prior \textit{supervised and unsupervised} methods on the NUCLA dataset. We improve by 8.5\% upon the previous RGB-based SOTA approach~\cite{viewclr}. 
On the PKU-MMD-II dataset, our method outperforms prior work, all of which are based on Pose modality, by 2.8\% as shown in Table~\ref{table:pku}. It is interesting to note that although the Pose modality shows superior performance in supervised setting (Table~\ref{table:ntu120}), it lags behind when used for self-supervised learning in both in-domain fine-tuning (Table~\ref{table:ntu120}) and transfer learning (Table~\ref{table:pku}) settings. Finally, we show results on the ROCOG-v2 dataset in Table~\ref{table:rocog}, where we gain an improvement of 2\%. 
These transfer learning results clearly demonstrates that the representations  learnt using our approach generalize well.

\subsection{Ablation Study and Analysis}

\begin{figure}[h!]
\begin{center}

   \includegraphics[scale=0.45]{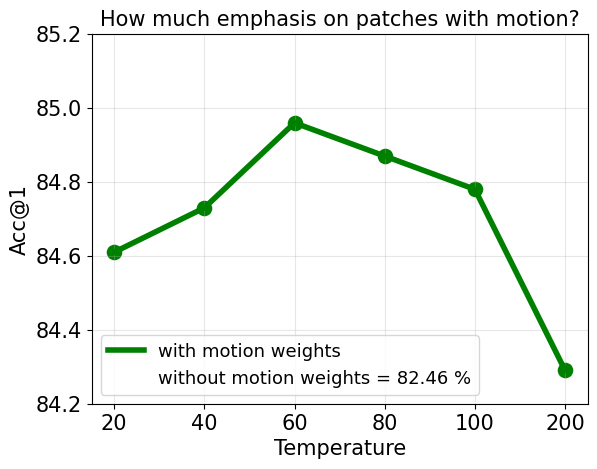}
   
\end{center}
\vspace{-10pt}
   \caption{\textbf{Temperature parameter} of motion weights modulates the focus on static vs moving regions as visualized in Figure~\ref{fig:motion-weights-examples}.}
\label{fig:ablation-temp}
\vspace{-10pt}
\end{figure}

\vspace{5pt}\noindent\textbf{How much emphasis to place on reconstructing moving patches?} In our approach, the motion weights can be adjusted to modulate the emphasis on moving patches using the temperature parameter in Algorithm~\ref{code}. As shown in Figure~\ref{fig:motion-weights-examples}, lower temperature value places more focus on reconstructing patches with more motion, and increasing the temperature increases the weight given to the background pixels.
Figure~\ref{fig:ablation-temp} shows the influence of the temperature parameter on accuracy. From the plot, we see that a temperature value of 60 performs best, which is used in all our experiments. Increasing the weights of background patches by increasing temperature degrades the performance. This is because it is trivial to reconstruct the background patches by copy-paste from nearby frames. The performance degrades significantly to 82.46\% if each patch is weighted equally.       

\vspace{5pt}\noindent\textbf{Masking Ratio.}
We study the impact of masking ratio in Table~\ref{table:ablation-maskratio}. We note that the optimal masking ratio is lower in our multi-view setting than the single-view setting in \cite{videomae}, which we hypothesize is because the model needs more information from each individual view to effectively infer cross-view geometry.
\begin{table}[ht]
\caption{\textbf{Masking Ratio.} MV2MAE performs best with a masking ratio of 0.7 which is needed for effectively inferring cross-view geometry.}
  \begin{center}
    {{
\begin{tabular}{lcccc}
\toprule
Masking ratio ($\rho$) & 0.6 & 0.7 & 0.8 & 0.9 \\
\midrule
NTU-120 xsub (\%) & 83.6 & \textbf{85.3} & 84.3 & 83.4 \\

\bottomrule
\end{tabular}
}}
\end{center}
\label{table:ablation-maskratio}
\end{table}

\vspace{5pt}\noindent\textbf{Model Scaling.}
To study how the performance scales with models of different capacities, we compare the fine-tuning performance of pre-trained models with ViT-T, ViT-S, and ViT-B in Table~\ref{table:ablation-modelscale} on the NTU-120 cross-subject setting. Our approach effectively pre-trains larger models using the same amount of data.
\begin{table}[ht]

\caption{\textbf{Increasing Model Capacity}. We observe that our approach scales effectively with bigger models.}
\vspace{-10pt}
  \begin{center}
    {{
\begin{tabular}{lccc}
\toprule
Backbone & ViT-T & ViT-S & ViT-B \\
\midrule
NTU-120 xsub (\%) & 82.0 & 83.4 & 85.1 \\
\bottomrule
\end{tabular}
}}
\end{center}
\label{table:ablation-modelscale}
\vspace{-10pt}
\end{table}

\begin{figure*}[t]
\begin{center}
   \includegraphics[scale=0.36]{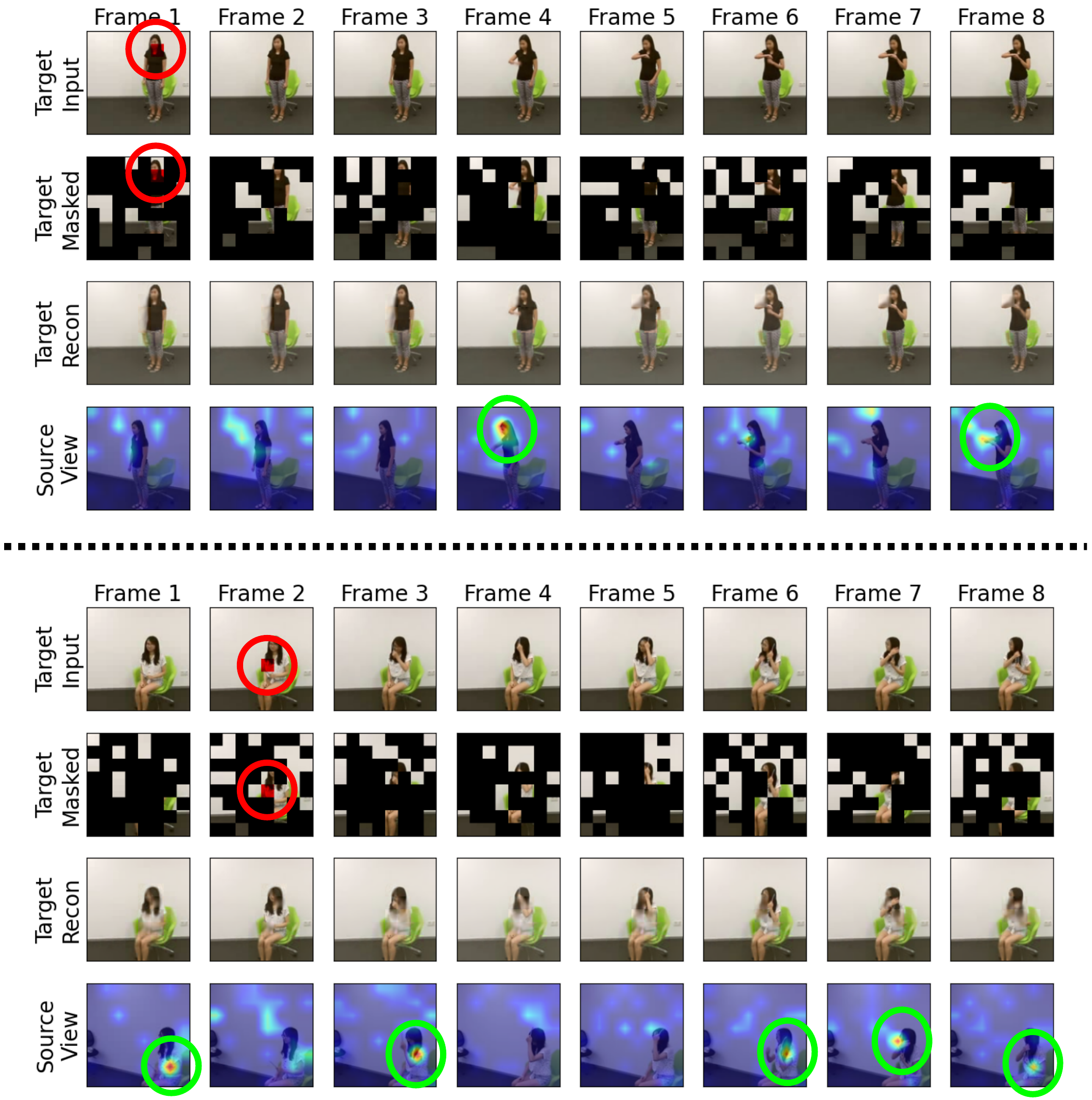}
\end{center}
    \vspace{-10pt}
    \caption{\textbf{Cross-View Decoder Qualitative Analysis.} We visualize the reconstructions and cross-attention maps from the cross-view decoder. \textbf{First row:} Target viewpoint input frames, \textbf{Second row:} Masked input frames from target viewpoint, \textbf{Third row:} Reconstruction of target view from the cross-decoder, \textbf{Last row:} Cross-attention maps visualized on source view frames. The \textcolor{red}{red} circle indicates the query token whose attention maps are visualized. \textcolor{green}{green} circles shows that the model is able find matching regions across viewpoints.}
\label{fig:xattn}
\end{figure*}

\vspace{5pt}\noindent\textbf{Visualizing Cross-Attention Maps and Reconstructions.}
Here, we analyze the cross-view decoder by visualizing the cross-attention maps and the cross-view reconstruction quality. The cross-attention maps are visualized in Figure~\ref{fig:xattn}. The first and second rows show the input and masked input frames from the target viewpoint, with the masked query token circled in red. The third row shows the reconstructed target view from the cross-view decoder. The last row shows the cross-attention map for the query overlaid on the source view frames. We can see that model is able to find matching regions in the source view, demonstrating the learnt geometry. 

\vspace{5pt}\noindent\textbf{How many source views to use?}
For the cross-view decoder, we study the effect of number of source viewpoints used in Table~\ref{table:ablation-num-views}. For these experiments, all viewpoints used are chosen randomly from available synchronized views. The performance is similar when using one or two source viewpoints. We observe that the fine-tuning accuracy drops if we use more source viewpoints for reconstructing the target viewpoint, by making the reconstruction task easier.

\begin{table}[ht]

\caption{\textbf{Number of Source Views.} Having more source views makes reconstruction task easier and degrades performance.}
\vspace{-10pt}
  \begin{center}
    {
\begin{tabular}{lccc}
\toprule
$\#$ Source Views & 1 & 2 & 3 \\
\midrule
ETRI xsub (\%) & 94.0 & 93.9 & 93.1 \\ 
\bottomrule
\end{tabular}
}
\end{center}
\label{table:ablation-num-views}
\end{table}

\begin{table}[ht]
\caption{\textbf{Which views to choose?} We study the effect of the distance between the source and target views. Keeping the target view fixed (View1), we vary the choice of source view and find that the performance degrades if the two views are very different from each other.}
\vspace{-10pt}

  \begin{center}
    {
\begin{tabular}{lccc}
\toprule
Source View & View2 & View3 & View4 \\
\midrule
ETRI xsub (\%) & 94.7 & 94.4 & 94.3 \\
\bottomrule
\end{tabular}
}
\end{center}
\label{table:ablation-which-views}
\vspace{-10pt}
\end{table}

\begin{figure}[t]
\begin{center}
   \includegraphics[scale=0.3]{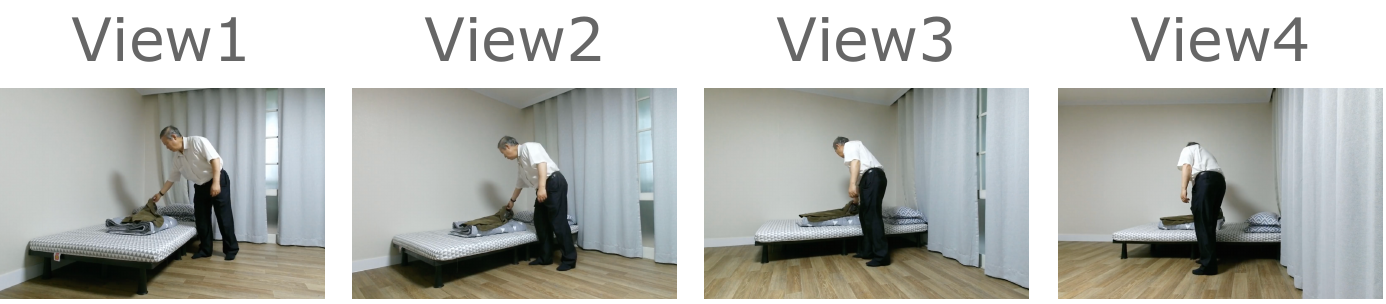}
\end{center}
    \vspace{-10pt}
    \caption{Example of synced views from ETRI dataset.}
\label{fig:views}
\end{figure}

\vspace{5pt}\noindent\textbf{How different should the views be?}
A natural question that arises is which viewpoint should we use? In other words, given a target viewpoint, how far should the source view be? We study this by fixing the target view to be View1, and varying the source view to be View2, View3 or View4, as shown in Figure~\ref{fig:views}. The results are reported in Table~\ref{table:ablation-which-views}, which shows that the performance drops if the chosen target and source viewpoints are separated by a lot.

\vspace{-10pt}

\section{Conclusion}

In conclusion, this paper proposes a self-supervised learning approach for harnessing the power of multi-view videos within the masked autoencoder framework. Our method integrates a cross-view reconstruction task, leveraging a dedicated decoder equipped with cross-attention mechanism to instill essential geometry information into the model. The introduction of a motion-focused reconstruction loss further enhances temporal modeling. Through comprehensive evaluation using full fine-tuning and transfer learning settings on multiple datasets, our approach consistently exhibits remarkable efficacy.

\clearpage
{
    \small
    \bibliographystyle{ieeenat_fullname}
    \bibliography{main}
}

\newpage
\section*{Appendix}
\appendix

\section{Architecture details}

The detailed asymmetric architecture of the encoder and decoders is shown in Table~\ref{tab:arch-enc}, Table~\ref{tab:arch-dec1} and Table~\ref{tab:arch-dec2}. We have two decoders in our architecture: 1) self-view decoder and 2) cross-view decoder. The self-view decoder only uses self-attention to reconstruct same view whereas the cross-view decoder uses cross-attention in addition to self-attention for reconstructing target viewpoint while also using source viewpoint. These decoders are discarded during fine-tuning. We use 16 frame input and choose ViT-S/16 as our default encoder. We adopt the joint space-time attention~\cite{vivit} for the encoder. 

\begin{table}[h]
    \scriptsize
    \centering
    \tablestyle{1pt}{1.08}
    \begin{tabular}{c|c|c}
    	\textbf{Stage} & \textbf{Vision Transformer (Small)} & \textbf{Output Sizes}  \\  
    	\shline
    	data & stride {4}\x\xycolor{1}\x\xycolor{1} &   \wcolor{3}\x\tcolor{16}\x\xycolor{128}\x\xycolor{128}  \\
        \hline
    	\multirow{2}{*}{cube} & \tcolor{2}\x\xycolor{16}\x\xycolor{16}, {\wcolor{384} } & \multirow{2}{*}{ \wcolor{384}\x\tcolor{\textbf{8}}\x\xycolor{64}} \\
    	& stride \tcolor{2}\x\xycolor{16}\x\xycolor{16}&  \\
    	\hline 
    	\multirow{2}{*}{mask} & random mask  & \multirow{2}{*}{ \wcolor{384}\x\tcolor{\textbf{8}}\x[\xycolor{64}\x(1-\maskcolor{$\rho$})]}\\ 
    	&  \emph{mask ratio = \maskcolor{$\rho$}} &  \\ 
    	\hline 
            \multirow{2}{*}{encoder} & \blockatt{384}{1536}{12} &  \multirow{2}{*}{ \wcolor{384}\x\tcolor{\textbf{\textbf{8}}}\x[\xycolor{64}\x(1-\maskcolor{$\rho$})]}\\
        & &  \\
        \hline
        \multirow{2}{*}{projector} & \text{MLP(\wcolor{192})} \& & \multirow{2}{*}{ \wcolor{192}\x\tcolor{\textbf{8}}\x\xycolor{64}} \\ 
        & \emph{concat learnable tokens} &  \\
    
	\end{tabular}
	\vspace{0.2em}
	\caption{\textbf{Encoder of MV2MAE.} The encoder processes 16-frame input clips from source and target views, and the encoded representations of the visible tokens are combined with the learnable mask tokens, before passing through the decoder.}
	\label{tab:arch-enc}
\end{table}

\begin{table}[h]
    \scriptsize
    \centering
    \tablestyle{1pt}{1.08}
    \begin{tabular}{c|c|c}
    	\textbf{Stage} & \textbf{Transformer} & \textbf{Output Sizes}  \\  
    	\shline
            \multirow{2}{*}{self-view decoder} & \blockatt{192}{768}{4} &   \multirow{2}{*}{ \wcolor{192}\x\tcolor{\textbf{8}}\x\xycolor{64}}\\
            & & \\
            \hline 
            projector & \text{MLP(\wcolor{1536})} &   \wcolor{1536}\x\tcolor{\textbf{8}}\x\xycolor{64} \\ 
            \hline
            reshape & \emph{from} \wcolor{1536} \emph{to}   \wcolor{3}\x\tcolor{2}\x\xycolor{16}\x\xycolor{16} &  \wcolor{3}\x\tcolor{16}\x\xycolor{128}\x\xycolor{128} \\
	\end{tabular}
	\vspace{0.2em}
	\caption{\textbf{Self-view decoder of MV2MAE.} It takes the source and target view tokens and reconstructs both the views independently.}
	\label{tab:arch-dec1}
\end{table}

\begin{table}[h]
    \scriptsize
    \centering
    \tablestyle{1pt}{1.08}
    \begin{tabular}{c|c|c}
    	\textbf{Stage} & \textbf{Transformer} & \textbf{Output Sizes}  \\  
    	\shline
            \multirow{3}{*}{cross-view decoder} & \blockxatt{192}{768}{4} &   \multirow{3}{*}{ \wcolor{192}\x\tcolor{\textbf{8}}\x\xycolor{64}}\\
            & & \\
            & & \\
            \hline 
            projector & \text{MLP(\wcolor{1536})} &   \wcolor{1536}\x\tcolor{\textbf{8}}\x\xycolor{64} \\ 
            \hline
            reshape & \emph{from} \wcolor{1536} \emph{to}   \wcolor{3}\x\tcolor{2}\x\xycolor{16}\x\xycolor{16} &  \wcolor{3}\x\tcolor{16}\x\xycolor{128}\x\xycolor{128} \\
	\end{tabular}
	\vspace{0.2em}
	\caption{\textbf{Cross-view decoder of MV2MAE.} The cross-view decoder uses the visible tokens from the source view to reconstruct the missing patches in the target view. This cross-view information is pulled in using the cross-attention block (MHCA).}
	\label{tab:arch-dec2}
\end{table}

\section{Implementation details}
The pre-training and fine-tuning hyper-parameter settings for NTU-60, NTU-120 and ETRI datasets are given in Table~\ref{tab:pt-setting} and Table~\ref{tab:ft-setting}. 

\begin{table}[t!]
\tablestyle{1pt}{1.02}
\small
\begin{tabular}{y{88}|x{58}x{58}x{58}}
config & NTU60 & NTU120 & ETRI \\
\shline
optimizer & \multicolumn{3}{c}{AdamW} \\ 
base learning rate & \multicolumn{3}{c}{1e-3}\\
weight decay & \multicolumn{3}{c}{0.05} \\
optimizer momentum & \multicolumn{3}{c}{$\beta_1, \beta_2{=}0.9, 0.95$} \\
batch size & \multicolumn{3}{c}{1024} \\
learning rate schedule & \multicolumn{3}{c}{cosine decay} \\
warmup epochs & 320 & 160 & 160 \\
total epochs & 3200 & 1600 & 1600 \\
augmentation & \multicolumn{3}{c}{MultiScaleCrop} \\
\end{tabular}
\vspace{.2em}
\caption{\textbf{Pre-training setting.}}
\label{tab:pt-setting} 
\end{table}
\begin{table}[t!]
\tablestyle{1pt}{1.02}
\small
\begin{tabular}{y{85}|x{58}x{58}x{58}}
config & NTU60 & NTU120 & ETRI \\
\shline
optimizer & \multicolumn{3}{c}{AdamW} \\
base learning rate & \multicolumn{3}{c}{1e-3} \\
weight decay & \multicolumn{3}{c}{0.1} \\
optimizer momentum & \multicolumn{3}{c}{$\beta_1, \beta_2{=}0.9, 0.999$} \\
batch size & \multicolumn{3}{c}{1024} \\
learning rate schedule & \multicolumn{3}{c}{cosine decay} \\
warmup epochs & 5 & 10 & 10 \\
training epochs &  35 & 120 & 120 \\
repeated augmentation &\multicolumn{3}{c}{6}  \\
flip augmentation & \multicolumn{3}{c}{\emph{yes}} \\
RandAug  & \multicolumn{3}{c}{(7, 0.5)} \\
label smoothing & \multicolumn{3}{c}{0.1} \\
drop path & \multicolumn{3}{c}{0.1} \\
layer-wise lr decay  & \multicolumn{3}{c}{0.9}  \\
\end{tabular}
\vspace{.2em}
\caption{\textbf{End-to-end fine-tuning setting}}
\label{tab:ft-setting}
\end{table}

\section{Comparison with VideoMAE}
VideoMAE~\cite{videomae} proposes to use the tube masking for dealing with the temporal redundancy in videos. Instead, we use a motion re-weighted reconstruction loss to deal with this issue. Table~\ref{table:vs-vmae} shows that our approach to tackle temporal redundancy is superior to using tube masking. 
\begin{table}[h!]

\vspace{-10pt}
  \begin{center}
    {{
\begin{tabular}{lc}
\toprule
Method & NTU-120 xsub (\%) \\
\midrule

Tube masking (VideoMAE) & 79.7 \\
Motion-weighted rec. loss (MV2MAE) & 84.8 \\

\bottomrule
\end{tabular}
}}
\end{center}
\caption{Motion-weighted reconstruction loss in MV2MAE is a more effective way of combating temporal redundancy in videos compared to tube masking of VideoMAE.}
\label{table:vs-vmae}
\end{table}

\section{Single-View vs Multi-View Inference}
At test time, multiple viewpoints of an activity are available in the cross-subject setting. However, evaluation in prior work is carried out using single-view at a time, following the original benchmark~\cite{ntu60}. Though in most practical scenarios, it would be natural to combine the predictions from available synchronized viewpoints for a given activity. We show this comparison of single-view and multi-view inference in Table~\ref{table:ablation-sv-mv-inf}. For multi-view inference, the predictions are combined using late fusion strategy. 
\begin{table}[h!]

  \begin{center}
    {{
\begin{tabular}{lcc}
\toprule
 & \multicolumn{2}{c}{Cross-Subject (\%)}\\
Method & NTU-60 & NTU-120 \\
\midrule

Single-View Inference & 90.0 & 85.3 \\
Multi-View Inference & 91.9 & 87.9 \\

\bottomrule
\end{tabular}
}}
\end{center}
\caption{SV vs MV inference.  We perform late fusion for multi-view inference.}

\label{table:ablation-sv-mv-inf}
\end{table}

\section{Synthetic data for pre-training}
\begin{figure}[t]
\begin{center}
   \includegraphics[scale=0.55]{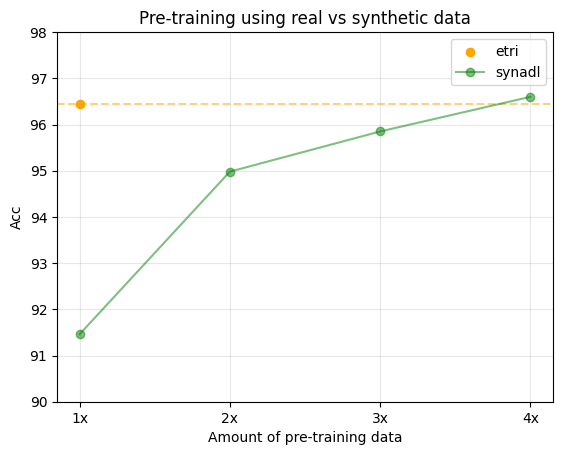}   
\end{center}
   \caption{\textbf{Pre-training using synthetic data.} Pre-training using more (4x) synthetic data beats pre-training using real data on the same real test set.}
\label{fig:ablation-synadl-amt}
\end{figure}

Real multi-view videos can be difficult to acquire and can pose privacy concerns. Here we explore using synthetic multi-view action recognition data as an alternative. In these experiments, the pre-training is done on synthetic data (SynADL~\cite{hwang2021eldersim} dataset) while fine-tuning and inference is done on the real data (ETRI~\cite{etri} dataset). We compare synthetic pre-training (\textcolor{ForestGreen}{green}) with real pre-training (\textcolor{orange}{orange}). We observe that if the amount of synthetic data used is same (1x) as the amount of real data, there is a performance drop due to the domain difference. Interestingly, if we increase the amount of synthetic data used for pre-training, synthetic pre-training can outperform real pre-training, as seen in Figure~\ref{fig:ablation-synadl-amt}. 

\end{document}